\documentclass[nohyperref]{article}

\usepackage[svgnames,dvipsnames,color,table]{xcolor}

\usepackage{microtype}
\usepackage{graphicx}
\usepackage{subfigure}
\usepackage{booktabs} 
\usepackage{multirow}
\usepackage{makecell}

\usepackage{hyperref}

\hypersetup{
  colorlinks   = true, 
  urlcolor     = RoyalBlue, 
  linkcolor    = RoyalBlue, 
  citecolor   = RoyalBlue 
}

\def\code#1{\texttt{#1}}

\usepackage[accepted]{icml2023}

\usepackage{amsmath}
\usepackage{amssymb}
\usepackage{amsfonts}
\usepackage{mathtools}
\usepackage{amsthm}

\usepackage{hhline}

\usepackage[capitalize,noabbrev]{cleveref}

\theoremstyle{plain}

\theoremstyle{definition}

\theoremstyle{remark}

\usepackage[textsize=tiny]{todonotes}

\icmltitlerunning{Git-Theta}

\begin{document}

\twocolumn[
\icmltitle{Git-Theta: A Git Extension for Collaborative Development of \\Machine Learning Models}



\icmlsetsymbol{equal}{*}

\begin{icmlauthorlist}
\icmlauthor{Nikhil Kandpal}{equal,unc}
\icmlauthor{Brian Lester}{equal,unc}
\icmlauthor{Mohammed Muqeeth}{unc}
\icmlauthor{Anisha Mascarenhas}{unc}
\icmlauthor{Monty Evans}{unc}
\icmlauthor{Vishal Baskaran}{unc}
\icmlauthor{Tenghao Huang}{unc}
\icmlauthor{Haokun Liu}{unc}
\icmlauthor{Colin Raffel}{unc}
\end{icmlauthorlist}

\icmlaffiliation{unc}{Department of Computer Science, University of North Carolina, Chapel Hill, USA}

\icmlcorrespondingauthor{Brian Lester}{bdlester@cs.unc.edu}

\icmlkeywords{Machine Learning, ICML, Version Control, Model Checkpoint, Git}

\vskip 0.3in
]



\printAffiliationsAndNotice{\icmlEqualContribution} 

\begin{abstract}
Currently, most machine learning models are trained by centralized teams and are rarely updated.
In contrast, open-source software development involves the iterative development of a shared artifact through distributed collaboration using a version control system.
In the interest of enabling collaborative and continual improvement of machine learning models \citep{raffel2023building}, we introduce Git-Theta, a version control system for machine learning models.
Git-Theta is an extension to Git, the most widely used version control software, that allows fine-grained tracking of changes to model parameters alongside code and other artifacts.
Unlike existing version control systems that treat a model checkpoint as a blob of data,
Git-Theta leverages the structure of checkpoints to support communication-efficient updates, automatic model merges, and meaningful reporting about the difference between two versions of a model.
In addition, Git-Theta includes a plug-in system that enables users to easily add support for new functionality.
In this paper, we introduce Git-Theta's design and features and include an example use-case of Git-Theta where a pre-trained model is continually adapted and modified.
We publicly release Git-Theta in hopes of kickstarting a new era of collaborative model development.\footnote{\url{https://github.com/r-three/git-theta/}}
\end{abstract}

\section{Introduction}

Over the past fifty years, software development has gone from a poorly understood and unregimented practice to a well-studied set of processes and methodologies that help ensure that software is performant, high-quality, and correctly implemented \citep{brooks1975mythical}.
A key component of modern software development is \textit{version control systems}, which allow fine-grained tracking of changes to a piece of software.
Typically, version control systems also enable collaborative development of software by providing functionality for soliciting and incorporating changes from contributors.
This functionality is key to the development of \textit{open-source} software, i.e., software whose source code is developed and released publicly.
Open-source software underlies many key technologies, including operating systems, compilers, web servers, programming languages, and more.
Much of this ubiquitous open-source software is developed by distributed communities of individuals who iteratively build, update, and improve the software through huge numbers of contributed changes.

In the last few decades, machine learning models have increasingly become a key part of software systems.
However, the development of machine learning models is relatively primitive in comparison with software development.
For example, in many applications of machine learning, it can be useful to update a trained model---e.g., to fix problematic behavior, add new capabilities, or learn from newly collected data.
While sophisticated tools such as DVC, MLFlow, and others (see \cref{sec:related-work} for a discussion) provide useful functionality for keeping track of subsequent versions of a machine learning model, these tools treat the machine learning model's checkpoint (the saved parameter values) as a blob of data; i.e., they are simply treated as a generic large file.
As such, a change to \emph{any} of the model's parameters incurs the same communication and storage costs as changing \emph{all} of the model's parameters.
As an additional consequence, these tools cannot support important operations involved in collaborative development such as merging independently introduced updates from different contributors.
We hypothesize that this lack of functionality is a major reason that machine learning models are almost never built collaboratively.
However, we are optimistic that large-scale collaboration is possible.
As a motivating example, there are currently over 10,000 variants of the BERT model on the Hugging Face Model Hub.\footnote{\url{https://huggingface.co/models?search=bert}}
What if the collective efforts used in these many training runs could be tracked, combined, and reused?

In this work, we aim to address these issues by introducing Git-Theta, a version control system that supports cheaply communicable updates as well as methods for merging updates from different contributors.
Git-Theta is built on top of recent research that has shown that a model can be updated by only training a relatively tiny number of parameters \citep{lester2021tuning,hu2021lora,sung2021sparse,guo2020diffpruning,liu2022tfew,he2022towards,ding2022delta} and that the capabilities of independently trained models can be combined into a single model without re-training \citep{matena2021merging,choshen2022fusing,wortsman2021robust}.
To ensure broad applicability and long-term relevance of Git-Theta, we include a plug-in system that makes it straightforward to add new update types, merging methods, and supported checkpoint formats.
We built Git-Theta as an extension of Git, the most widely used version control system (according to the StackOverflow 2022 Developer Survey,\footnote{\url{https://survey.stackoverflow.co/2022/\#version-control-version-control-system-prof}} over 95\% of professional developers primarily use Git).
As such, Git-Theta can be easily integrated into existing software development pipelines and allows tracking both the code and parameters of a model in the same repository.
Compared to existing systems for tracking models that treat the checkpoint as a generic blob of data, Git-Theta views model checkpoints as a collection of ``parameter group'' tensors that can be modified through communication-efficient updates and merged by manipulating their values.
This approach allows core Git concepts to be applied at the parameter level, making Git-Theta dramatically more communication- and space-efficient while facilitating automatic merging and providing meaningful difference information when models have changed.
We are optimistic that these features will provide a major step forward in enabling collaborative and continual development of machine learning models.

\section{Background} \label{sec:background}

In this section, we provide a brief introduction to Git, the version control system upon which Git-Theta is based.
In particular, we focus on Git's built-in support for extensions like Git-Theta.
We also discuss Git LFS (a Git extension that allows Git to better handle large files) as Git-Theta uses Git LFS for parameter storage.

\subsection{Version Control with Git} \label{ssec:git}

Git is an open-source version control system that was created in 2005 to support collaborative development of the Linux kernel. Since then, Git has become one of the most widely used version control systems for both open- and closed-source software.

In a typical Git workflow, the collection of code corresponding to a piece of software is hosted on the \textit{main branch}---i.e., the sequence of code snapshots that correspond to the canonical version of the software---on a Git remote server (common commercial examples of Git remote hosts include GitHub, GitLab, and BitBucket).
Contributors independently propose changes to the main branch by downloading a full copy of the repository from the remote (\code{git fetch}), committing changes locally (\code{git commit}), and sending their changes back to the remote repository (\code{git push}).

In typical workflows, contributors to a Git repository will create a new local branch when developing changes (\code{git~branch}). Creating a branch allows one contributor to make a series of commits in isolation from any concurrent changes that other contributors make on the main branch or other branches. Additionally, branching makes switching between logically separate features easy (\code{git~checkout}). Once development on a branch is done, the commits from the branch can be incorporated into the main branch (\code{git~merge}), adding a new snapshot to the main branch's history. Additionally, Git provides several conflict resolution methods for merging branches that contain overlapping changes to the same files. 

Internally, Git manages repositories by storing files in three logically separate locations. First is the \emph{working tree}, where files in a repository are viewed and edited by users. The second location is the \emph{staging area}, a location managed internally by Git that contains modified files that are ready to be committed. Git places files in the staging area when users run \code{git add}. Finally, after staged files are committed and a user runs \code{git push}, the committed files are stored on the \emph{Git remote}. Other developers can then retrieve these changes via \code{git pull}.

\subsection{Efficient Storage in Git} \label{ssec:git-storage}

Git stores the history of a repository as a sequence of snapshots. In each snapshot, the state of each file in the repository is recorded. To do this efficiently, Git only records files that have changed; 
unchanged files simply point to the last snapshot in which they were modified.

When version-controlled files are small, this approach works well. However, when a file is large and is changed often, more and more copies are saved and the repository becomes bloated. The inefficiencies of this approach are magnified when only small parts of the large file are changed as large swaths of duplicated data get saved.
To overcome this issue when tracking large model checkpoint files, Git-Theta leverages the structure of model checkpoints to extend the idea of snapshots to individual parameter groups (described further in \cref{sec:git_theta}).

\subsection{Extending Git} \label{ssec:extending_git}
Git offers several points of customization that allow external programs to extend its functionality to fit the needs of a new use-case. Extensions can be created based on per-file customizations and repository-level event (e.g., \code{commit}, \code{merge}, etc.) customizations.

\begin{figure*}[!t]
  \begin{center}
    \centerline{\includegraphics[width=0.8\textwidth]{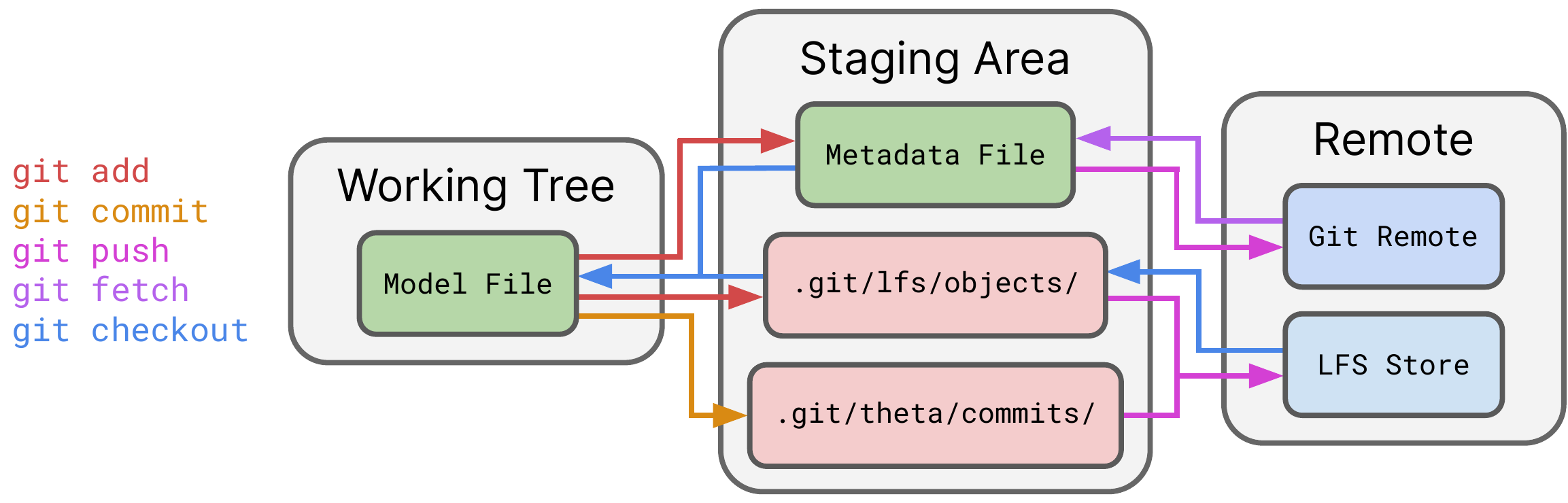}}
    \caption{An overview of how Git-Theta manages a model checkpoint file during the five basic Git operations. The flow of data between the working tree, staging area, and remote associated with each Git command (left) is shown by the arrows matching the command's color. For simplicity we include \code{.git/lfs/objects/} and \code{.git/theta/commits/} in the staging area since they are \emph{internal} data structures used by Git-Theta and Git LFS.}
    \label{fig:git-flow}
  \end{center}
  \vskip -0.4in
\end{figure*}

\paragraph{Git Attributes: Per-File Customization}
At specific points in the Git workflow (see \cref{fig:git-flow}) Git is able to call external programs to customize how it handles specific files. 
These customizations are configured via the \code{.gitattributes} file in the repository. Each entry in this file contains a glob pattern that specifies which files it applies to as well as a sequence of key-value pairs that represent attributes of these files. These attributes dictate how Git handles those files, including calling external programs for certain operations.

\paragraph{The Filter Attribute}
The \code{filter} attribute customizes how files are moved between the working tree and staging area.
When staging a working tree file (i.e., running \code{git add}), Git passes the file to the \emph{clean filter} associated with the value of that file's \code{filter} attribute. The clean filter is a program that can perform arbitrary transformations to the working tree file, and the output of the filter is what Git places in the staging area and tracks in version control.
Conversely, when the file is moved from the staging area back to the working tree, Git passes the staged file to the \emph{smudge filter} associated with the file's \code{filter} attribute. The smudge filter is typically the inverse of the clean filter, resulting in the original contents of the file being repopulated in the working tree.
Together, the clean and smudge filters allow files to be represented differently in the working tree and the underlying version control system.

\paragraph{The Diff Attribute}
By default, Git uses its built-in text-based diff driver to compute and display differences between two versions of a file. However, this can be customized via the \code{diff} attribute, causing \code{git diff} to call an external diff program to compute and display differences.

\paragraph{The Merge Attribute}
Git attempts to perform automatic merging of files from different branches during the \code{git merge} operation using its built-in text-based merge driver. The \code{merge} attribute can be used to specify an external merging program to use instead.

\paragraph{Git Hooks: Repository-level Customization}
At specific points in the Git workflow, Git can invoke external programs, called hooks, that operate on the whole repository. For example, when a user runs \code{git commit}, Git first invokes the \emph{pre-commit} hook. After completing the commit operation, Git then calls the \emph{post-commit} hook. Analogous \emph{pre-} and \emph{post-} hooks exist for other operations like \code{push} and \code{pull}. These allow external programs to inspect the files being manipulated by Git and perform additional operations alongside the ones Git performs itself.

\subsection{Large File Storage with Git LFS} \label{ssec:git_lfs}
Git was designed to track small text files such as source code. Git's limited tooling for dealing with binary files and its distributed, full-copy nature make it inappropriate for dealing with large binary files. Git Large File Storage (LFS)\footnote{\url{https://git-lfs.com/}} is a Git extension that makes Git more suitable for tracking large binary files.
At a high-level, Git LFS replaces large binary files in a repository with small text files that Git is better equipped to track. These small text files contain metadata about the file and where its contents are stored. Since Git-Theta leverages Git LFS to store the values of parameter group tensors and its design mirrors Git LFS, this section provides a brief overview of how Git LFS works.

\paragraph{Tracking Large Files}
A large file is tracked by Git LFS with the \code{git lfs track <file>} command. This creates a new entry in the \code{.gitattributes} file that sets the tracked file's \code{filter}, \code{diff}, and \code{merge} attributes. Additionally, Git LFS registers a pre-push hook. We describe these filters and the hook below.

\paragraph{Staging Large Files}
When an LFS-tracked file is staged via \code{git add}, Git passes the file to the LFS clean filter. This filter copies the working tree file to \code{.git/lfs/objects/} and returns a \emph{pointer file} containing the LFS-tracked file's hash and size as well as the current version of the LFS specification. Git then stores this pointer file in the staging area. Rather than tracking the version history of the large binary file, Git itself then only handles the small, text-based pointer file.

\paragraph{Pushing Large Files to a Remote}
Before Git pushes a sequence of commits to a remote repository, it invokes the LFS pre-push hook. This hook inspects each commit to see if it contains LFS pointer files. If any pointer files are encountered, the corresponding files in \code{.git/lfs/objects} are synced to an LFS remote server. This server is similar to a Git remote, but only stores LFS-tracked files. 
Once the pre-push hook has completed, Git pushes the commits containing pointer files to the Git remote. As a result, LFS-tracked files appear as pointer files when viewed in the remote repository, while the files' actual contents are stored on an LFS remote server.

\paragraph{Fetching Large Files from a Remote}
Git performs the \code{git fetch} operation as it would for any repository, regardless of whether it contains LFS-tracked files. However, since the Git remote contains LFS pointer files in place of the LFS-tracked files' actual contents, the result of a fetch is that only the pointer files are downloaded and placed in the staging area.

\paragraph{Checking Out Large Files}
When LFS-tracked files are moved from the staging area into the working directory, Git calls the LFS smudge filter. The smudge filter takes a staged pointer file as input and checks if the corresponding large file is cached locally in \code{.git/lfs/objects/}. If the file exists locally, it is returned by the smudge filter. In cases when the file is not cached locally, the smudge filter first retrieves the file from the LFS remote server. Git then populates the working tree with the true file contents returned by the smudge filter.

\section{Git-Theta} \label{sec:git_theta}
\subsection{Motivation}

Due to the difficulties of managing model checkpoints, researchers often use Git to track only what is required to recreate the model, i.e., the code that implements the model, downloads and processes the training dataset, and specifies the training configuration (e.g., hyperparameters). This approach makes it difficult to reconstruct the model's parameter values, which would require re-training and/or reconstructing actual changes to the model (e.g., adding or removing parameters) over time. It would thus be advantageous to be able to additionally track the model parameters in version control.

While Git LFS \textit{could} be used to track the version history of a model---in fact, many existing model versioning tools use Git LFS directly, see \cref{sec:related-work}---it is a general-purpose tool for working with large files and cannot take advantage of the structure of model checkpoints. This presents a number of usability issues when tracking a model's version history. First, when a model is tracked by Git LFS, it is treated as a blob of data that is versioned via Git's per-file, all-or-nothing snapshot approach. Thus \emph{any} change to a model file results in a new copy of the \emph{entire} model being stored. As a result, the storage requirements of the repository scale linearly with the number of model commits, independent of the number of parameters that are actually updated or how they are updated. This is especially problematic when using communication-efficient training methods that, e.g., update a small subset of the parameters \cite{sung2021sparse,guo2020diffpruning,zaken2021bitfit}, add a small number of new parameters \cite{liu2022tfew,hu2021lora,houlsby2019adapter,lester2021tuning,li2021prefix}, or optimize the model in a low-dimensional subspace \cite{aghajanyan2020intrinsic}.
An additional downside of treating model checkpoints as generic large files is that Git LFS cannot automatically merge files from different branches. Instead, Git LFS flags the files as having a merge conflict and defers to users to manually merge them. Due to its generality, Git LFS cannot not take advantage of existing methods for merging multiple models into a single model that combines their capabilities (e.g., \citealp{matena2021merging,wortsman2021robust,choshen2022fusing}).
Finally, being agnostic to the underlying structure of a tracked file, Git LFS cannot provide meaningful ``diffs'' between versions of a model. Git LFS can only indicate that two model checkpoints are not bitwise identical.

To avoid these issues, Git-Theta focuses specifically on tracking machine learning models and leverages the structure in model checkpoints to alleviate the aforementioned deficiencies of Git and Git LFS. By considering the ``parameter groups'' (e.g., weight matrices or bias vectors in neural networks) that make up a checkpoint, Git-Theta offers increased storage efficiency by applying the idea of snapshots at the parameter group level. When a model checkpoint is tracked, each parameter group is checked for a change and only new values are stored; unchanged values are replaced with references to their previous values. Additionally, when two checkpoints differ, Git-Theta provides much more meaningful diff information such as which parameter groups in the model have actually changed. Similarly, when a model has multiple different histories (e.g., from different contributors), Git-Theta can ignore parameter groups that are equivalent across histories and automatically apply user-specified merge operations to the ones that differ. As further discussed below, these characteristics enable Git-Theta to meaningfully and efficiently track models natively with Git.

\subsection{Lifecycle of a Git-Theta Model}
To provide a practical picture of Git-Theta's usage and implementation, we describe the lifecycle of a model checkpoint tracked by Git-Theta. Exact details of the implementation are expanded upon in \cref{sec:implementation}.

\paragraph{Tracking a Model}
Git-Theta begins tracking a model checkpoint with the command \code{git theta track <checkpoint>} which configures Git to call Git-Theta when appropriate by setting the file's \code{filter}, \code{diff}, and \code{merge} attributes in the \code{.gitattributes} file. 

\paragraph{Staging a Model}
Running \code{git add <checkpoint>} stages the model and triggers the Git-Theta clean filter. Internally, the clean filter uses a \code{Checkpoint} class to load the framework-native checkpoint into a standardized format. It then checks if a previous version of the model exists and identifies the parameter groups that have been added, removed, or modified. Since a parameter could erroneously appear to be modified due to floating point imprecision, parameter groups are compared through locality sensitive hashes (described in \cref{sec:implementation}).

Any parameter groups that were modified or added are then passed to an \code{Update} class that replaces the parameter group tensor with the smallest amount of information needed to describe how the parameter group was modified. For a full dense update of a parameter group, this results in just keeping the new parameter values. However, for communication-efficient updates, this can amount to storing significantly less information -- for example, a sparse update \citep{sung2021sparse,guo2020diffpruning} to a parameter group can be represented by the indices of the values that changed and the changed values.

The \code{Update} class then calls a \code{Serializer} to serialize the parameter group update as a blob of data. This serialized update is passed to the Git LFS clean filter, which stores the update in \code{.git/lfs/objects/} and returns the pointer file metadata associated with the serialized object.

Finally, after processing each parameter group, the Git-Theta clean filter creates a model metadata file. This file contains per-parameter group metadata such as information about each group's tensor (i.e., its shape, dtype, and locality sensitive hash), the metadata produced by the LFS clean filter, and additional information such as the update type (e.g., dense, sparse, low-rank, etc.). The text-based metadata file is what is actually staged and versioned by Git itself.

\paragraph{Committing a Model}
After committing a model with \code{git commit}, Git invokes the Git-Theta post-commit hook. The hook inspects the most recent commit, looking for Git-Theta model metadata files. When found, the hook inspects the metadata file, looking for parameter groups that were modified in that commit and thus written to \code{.git/lfs/objects} during staging. The hook then stores a list of these LFS objects in \code{.git/theta/commits/<commit\ hash>}.

\paragraph{Pushing a Model to a Remote}
When Git prepares to push a series of commits to a remote, Git invokes the Git-Theta pre-push hook. This hook uses the data in \code{.git/theta/commits/} to find parameter groups that were added or modified during the commits being pushed. The LFS objects associated with these parameters are synced to an LFS remote server using \code{git lfs push}.

\paragraph{Fetching a Model from a Remote}
Git performs the \code{git fetch} operation normally. However, since the Git remote contains a Git-Theta metadata file in place of the actual model, the result of a fetch is that only the metadata file is downloaded and placed in the staging area.

\paragraph{Checking Out a Model}
During operations like \code{git checkout}, the Git-Theta smudge filter is invoked as Git moves the metadata file from the staging area to the working tree. 
For each parameter group, the smudge filter uses the \code{Update} class to load parameter values asynchronously. To do so, the \code{Update} reads the metadata for a parameter group and passes the LFS metadata to the Git LFS smudge filter. Git LFS then retrieves the serialized update from either the local cache in \code{.git/lfs/objects} or the LFS remote server. The \code{Serializer} is used to convert the serialized update back into parameter values. If the update being loaded depends on previous parameters (e.g., a low-rank update that is applied on top of previous parameters), the \code{Update} traverses the repository's history and recursively loads previous versions of the parameter group until it can fully reconstruct the current value of the parameters.
After each parameter group has been loaded, the Git-Theta smudge filter uses the \code{Checkpoint} class to convert the collection of parameter groups back into a framework-native checkpoint. Finally, Git places this reconstituted checkpoint in the working tree.

\paragraph{Merging Models From Different Branches}
When merging commits that have made modifications to the same model checkpoint, Git-Theta's custom merge driver is used. The driver loads the model metadata file from both branches as well as their common ancestor. When merge conflicts are detected, a menu is shown to the user that lists possible merge resolutions (discussed below). The \code{Merge} that the user selects merges the parameters from the commits as appropriate and the resulting parameters are placed in the merged model. Finally, the merged model is output by the Git-Theta merge driver so that it can be committed by Git.

\paragraph{Diffing Models}
When a user runs \code{git diff} to display the differences between two versions of a model checkpoint, Git calls the Git-Theta diff driver. The driver identifies and displays which parameter groups were added, modified, and removed between the two checkpoint versions.

\subsection{Implementation} \label{sec:implementation}
Git is highly customizable due to its implementation of Inversion of Control \cite{johnson1988designing} where users design custom modules but Git controls when they are executed. In an effort to make Git-Theta similarly flexible, we also use this principle. 
Sepcifically, control flow for core functionality (such as filters, merge/diff drivers, and hooks) are defined by Git-Theta, but the details of how model checkpoints are loaded, serialized, stored, and merged are all handled by plug-ins. Plug-ins can easily be registered by users to customize Git-Theta using Python's \cite{van1995python} entry-point-based plug-in system. While Git-Theta ships with plug-ins---described below---that support many common frameworks and usage patterns, end-users can define their own to handle additional setups.

\paragraph{Checkpoints}
The first type of plug-in Git-Theta supports is the \code{Checkpoint} plug-in. \code{Checkpoints} are responsible for loading a framework-native checkpoint file into a standardized format in memory, identifying parameter groups, and saving in-memory models back onto disk in the same framework-native format. Git-Theta currently has built-in support for Pytorch \cite{paszke2019pytorch}, Tensorflow \cite{abadi2016tensorflow}, and Flax \cite{flax2020github} checkpoint formats. User-defined plug-ins allow for straightforward addition of new checkpoint formats or changing the logic for identifying parameter groups.

\paragraph{Updates}
The \code{Update} plug-in enables Git-Theta to store updates to a parameter group as efficiently as possible. Each \code{Update} plug-in supports a different update type to a parameter group (e.g., dense, sparse, low-rank, etc.). An \code{Update} can infer the update made given the previous value of the parameter group and extract the minimum amount of information needed to describe the update. For instance, the sparse \code{Update} plug-in computes the difference between two versions of a parameter group and extracts the coordinates and values of the non-zero elements. However, this feature can introduce numerical noise due to non-deterministic floating-point operations needed to compute the update \citep{goldberg1991every}. \code{Update} plug-ins therefore also support loading the values for an update using an external file with the same structure as the model checkpoint. This approach prevents a mismatch between the parameter-efficient values and the frozen model they were trained with (which can result in poor performance \citealp{lester-etal-2022-recycling}) by combining the original checkpoint and update in one checkpoint.

Additionally, an \code{Update} is also responsible for translating the information describing an update into actual parameter values. For update types that rely on previous versions of a parameter group (e.g., a sparse update is applied on top of existing parameter group value), the \code{Update} plug-in recursively loads previous versions of the parameter group from Git history until it can fully reconstruct the current value of the parameters. 

Git-Theta currently supports dense, sparse \cite{sung2021sparse,guo2020diffpruning}, low-rank \cite{hu2021lora}, and (IA)$^3$ \cite{liu2022tfew} updates to parameter groups.

\paragraph{Serialization} 
The \code{Serializer} plug-in is responsible for serializing the tensors produced by an \code{Update} for storage on disk. Currently Git-Theta supports a Tensorstore \cite{tensorstore2020} \code{Serializer}.
Some updates require storing multiple values, e.g., sparse updates store both the sparse values and the index of those values. In these cases, the serialized values are combined using \code{msgpack}.\footnote{\url{https://msgpack.org}} The resulting blob is what is eventually stored in Git LFS.

\paragraph{Storage}
Git-Theta uses Git LFS to store serialized parameters instead of storing them directly in Git. Once a parameter group has been serialized, it is passed via inter-process communication to the Git LFS clean filter, which stores the blob and returns a metadata pointer file. This Git LFS metadata is then stored in the Git-Theta model metadata file that is ultimately tracked by Git.
Similarly, when a parameter group needs to be smudged, Git-Theta sends that parameter group's metadata to the Git LFS smudge filter and retrieves the serialized parameter group. This allows Git-Theta to leverage Git LFS network efficiency features. For example, when running \code{git clone}, only the current commit's version of the parameters are downloaded from the remote. Additionally, downloaded parameter groups are automatically cached and reused when moving through Git history.

\paragraph{Merges}
The \code{Merge} plug-ins represent different strategies for combining two versions of the same parameter group from different branches. In the event that two branches modify the same parameter group and must be merged, the Git-Theta merge driver dynamically builds a menu of merge strategies based on the registered \code{Merge} plug-ins. Each \code{Merge} provides a summary of what it does, the keyword used to select its strategy, and what kinds of merge conflicts it is able to resolve, allowing the driver to build a menu with only relevant plug-ins.

Git-Theta currently supports \code{Merge} plug-ins that resolve merge conflicts by using the change from the current branch, using the change from the other branch, using the change from their common ancestor (i.e., throwing away both changes), or averaging the parameters from each branch \citep{wortsman2021robust,choshen2022fusing}.

\paragraph{Locality Sensitive Hash} \label{sec:LSH}
When a parameter group in a model remains unchanged from the previous commit, Git-Theta avoids storing a new copy of that parameter group. To detect if a given parameter group has changed without loading both the current and previous versions of the parameter group into memory, Git-Theta first compares the metadata associated with each version. Mismatches in metadata such as parameter shape or dtype immediately signal that a parameter group has changed. However, in cases where the content (i.e., the parameter values) of the parameter group has changed, Git-Theta compares the hashes of the parameter groups.

However, na\"{i}vely hashing the underlying bytes of a tensor is unreliable as a change in a single bit, caused by differences in machines or library versions, results in a completely different hash value. Additionally, when using incremental \code{Updates}, where the final parameter value is computed on the fly, numerical noise from different linear algebra implementations or parallelism can introduce tiny differences in parameter values. These errors are small enough that they typically do not effect results when used in machine learning models \cite{Gupta2015,Courbariaux2014}, but nevertheless result in bit-based hash mismatches.


To avoid this issue, Git-Theta hashes parameters using a locality sensitive hash (LSH) \cite{gim-99}. An LSH will hash similar items to the same same value and provide probabilistic bounds on false positives. Git-Theta uses an LSH that approximates Euclidean distance based on bucketed projection of each point onto a set of random lines \cite{Datar2004}. Additionally, the random pool approach from \citet{van-durme-lall-2010-online} is used to allow the LSH to hash weights of variable size. The Euclidean LSH is $(d_1, d_2, p_1=1-d_1, p_2=1-d_2)$ sensitive, that is, if two items $x$ and $y$ have a $\text{distance}(x, y) \leq d_1$ then the probability of the two items having the same hash is $\Pr[h(x) = h(y)] \geq p_1$ while values with a $\text{distance}(x, y) \geq d_2$ have colliding hashes with a probability of $\Pr[h(x) = h(y)] \leq p_1$. 

Git-Theta's LSH uses $16$ hash functions and is calibrated so that when two parameters have a Euclidean distance less than $1e^{-8}$ then they will have the same hash with a probability of at least $99$\%. As a safety precaution, weights that have a Euclidean distance $\in [1e^{-8}, 1e^{-6}]$ are checked with \code{np.allclose} before deciding if they match or not.

Note that parameter hashing and the LSH provide probabilistic bounds on detecting changes in the parameter space due to noise.
This does not guarantee that a given LSH-equivalent change in parameters will not result in a meaningful change in the model's predictions.
However, testing for changes in model's output would be intractible as it would require evaluation of the model over the whole data distribution in the worst case.

\section{Benchmarking} \label{sec:benchmarking}

To test the effectiveness of Git-Theta's approach to model versioning, we compare it with Git LFS in a representative community development workflow.
We start with T0 3B \cite{sanh2021multitask}, a pre-trained language model.
T0 3B is itself derived from a chain of models, starting with T5 1.1 XL \footnote{\url{https://github.com/google-research/text-to-text-transfer-transformer/blob/main/released_checkpoints.md\#t511}}, a language model trained using a span corruption objective on the C4 dataset \cite{2020t5}. To produce T0 3B, T5 1.1 XL was then adapted through further training on the prefix-LM task \citet{lester2021tuning} before finally being trained on a multitask mixture of prompted datasets to improve generalization on novel tasks \citep{sanh2021multitask}.

\begin{figure}[!t]
\begin{center}
    \centerline{\includegraphics[width=0.8\columnwidth]{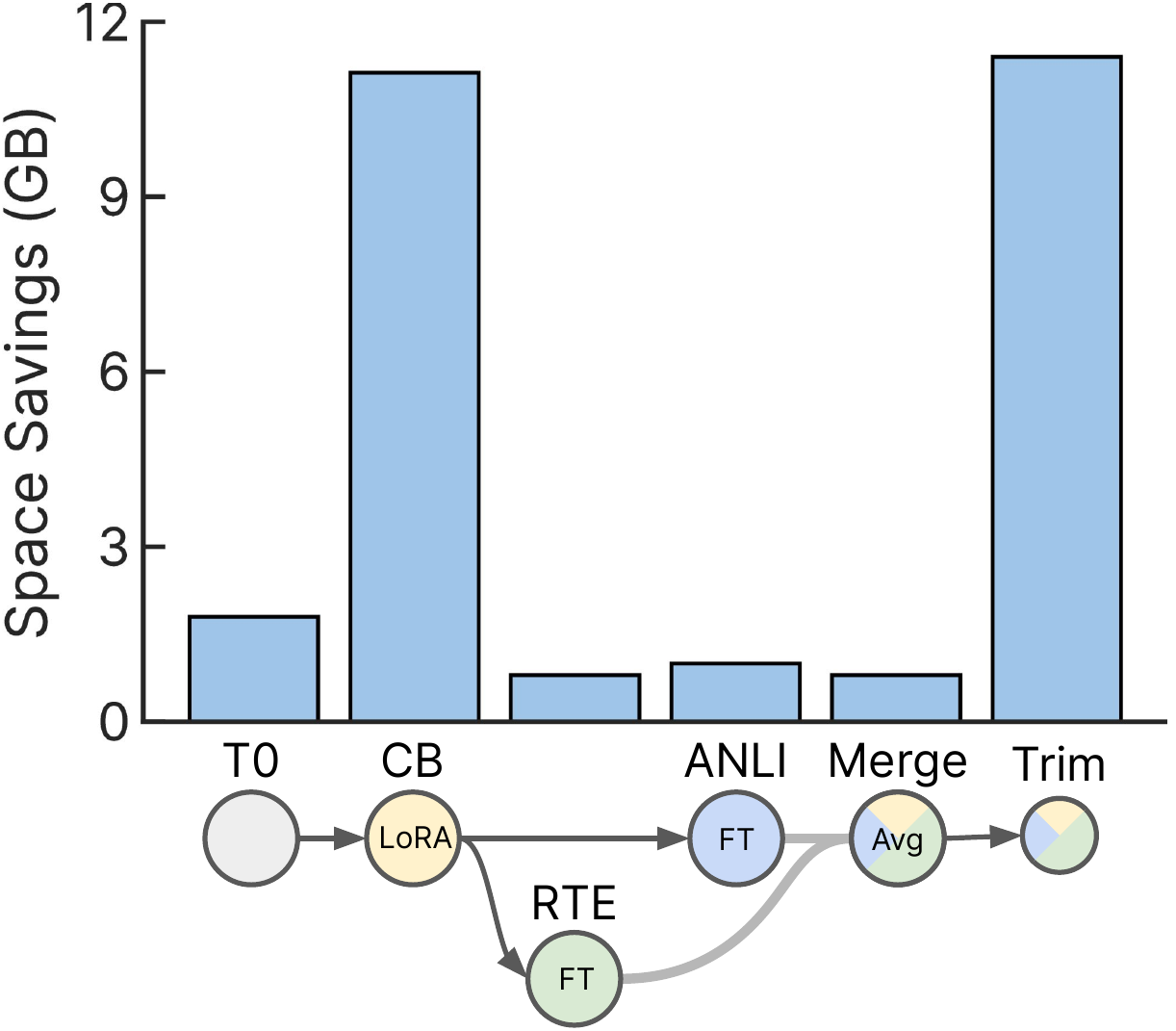}}
    \caption{Relative space saving of Git-Theta over Git LFS. Space savings are most apparent when using parameter-efficient training methods. Even when fine-tuning, where all parameters are updated and therefore need to be saved, Git-Theta still reduces storage requirements thanks to the way TensorStore compresses data.}
    \label{fig:storage}
\end{center}
\vskip -0.4in
\end{figure}

In our experiments, we first perform few-shot training of T0 3B on the CB dataset \cite{marneff-simons-tonhauser-2019} using low-rank updates (LoRA) \cite{hu2021lora} for a subset of the model's parameter groups.
On a new branch, the model is then few-shot fine-tuned on the RTE \cite{dagan2005pascal,bar2006second,giampiccolo2007third,bentivogli2009fifth} dataset. Concurrently, few-shot training on the ANLI R1 dataset \cite{nie2019adversarial} is performed on the main branch. The RTE-trained model is then merged back into the main branch using parameter averaging \cite{wortsman2021robust,choshen2022fusing}. Finally, we remove (Trim) the sentinel token embeddings used during pre-training of T5 \cite{2020t5} because they are not used after pre-training. Training hyperparameters follow those used in \citet{liu2022tfew}. \cref{fig:storage} (bottom) provides an overview of this development workflow.

\begin{table}[!t]
    \centering
    \footnotesize
    \begin{tabular}{l l r r}
      \toprule
      \textbf{Commit} & \textbf{Metric} & \textbf{Git LFS} & \textbf{Git-Theta} \\
      \midrule
      \multirow{3}{*}{\makecell[l]{Add T0 3B}} & \code{add}      & 2m 24.6s & 14m 24.4s \\
                       & \code{checkout} & 1m 9.5s  & 2m 33.1s \\
                       & Size     & 11.4GB   & 9.6GB \\
      \midrule
      \multirow{3}{*}{\makecell[l]{Train on CB with \\ LoRA}} & \code{add}      & 2m 21.6s & 7m 46.8s \\
                       & \code{checkout} & 35.6s    & 1m 3.3s \\
                       & Size     & 11.4GB   & 0.27GB \\
      \midrule
      \multirow{3}{*}{\makecell[l]{Fine-Tune on RTE}} & \code{add}      & 2m 24.1s & 13m 20.8s\\
                       & \code{checkout} & 1m 6.7s  & 2m 11.0s \\
                       & Size     & 11.4GB   & 10.62GB\\
      \midrule
      \multirow{3}{*}{\makecell[l]{Fine-Tune on ANLI}} & \code{add}      & 2m 23.0s & 10m 35.2s\\
                       & \code{checkout} & 1m 9.4s  & 1m 14s\\
                       & Size     & 11.4GB   & 10.4GB \\
      \midrule
      \multirow{3}{*}{\makecell[l]{Merge by averaging \\ parameters}} & \code{add}      & 2m 22.9s & 15m 51.2s\\
                       & \code{checkout} & 32.8s    & 1m 43.7s\\
                       & Size     & 11.4GB   & 10.6GB\\
      \midrule
      \multirow{3}{*}{\makecell[l]{Remove sentinels}} & \code{add}      & 2m 23.8s & 3m 41.3s \\
                       & \code{checkout} &  1m 6.5s & 1m 17.1s \\
                       & Size     & 11.4GB   & $10^{-5}$GB\\
      \midrule
      Total            & Size     & 57.0GB & 41.5GB \\
      \bottomrule
    \end{tabular}
    \caption{Comparison of speed and storage costs of Git LFS and Git-Theta. Git LFS is generally faster than Git-Theta, but Git-Theta can dramatically reduce storage and communication costs.}
    \label{tab:benchmark}
    \vskip -0.2in
\end{table}

After each training session, the model is added and committed to two repositories, one using Git LFS for version control and the other using Git-Theta. Since many model version control system either leverage Git LFS directly or use a similar approach (i.e., each version of the checkpoint is a new blob, see \cref{sec:related-work} for a discussion), we only compare to Git LFS.

At each commit we compare the wall-clock time for a \code{git add}, measuring how long the clean filter takes to run; the wall-clock time for a \code{git checkout}, measuring how long the smudge filter takes to run; and the size of the objects stored on disk. Data transferred to and from the remote is also an important metric for real-world utility, but eventually all checkpoint files need to be pushed to the remote so the on-disk size is a good proxy for the amount of data sent over the network. Git LFS, and by extension Git-Theta, use file hashing to avoid sending redundant data over the wire. As Git-Theta stores each parameter group independently, there may be small additional network overhead from creating additional TCP connections, but we expect that the increased parallelism and decrease in the total amount of data moved to more than compensate for any associated delays. Transfer times are strongly correlated to file size and differences would mostly amount to variance in the network, therefore we omit transfer times here.

All benchmarks were run on an Intel(R) Xeon(R) Silver 4214R CPU @ 2.40GHz with 48 cores and 256 GB of DDR4 RAM running at 2400 MT/s. Checkpoints were saved to a SATA SSD. 

\paragraph{Results}
\cref{tab:benchmark} shows the speed and storage---and thus communication---costs of each version control system; \cref{fig:storage} illustrates the relative storage savings at each step from using Git-Theta instead of Git LFS.
In general, Git LFS is consistent in terms of the wall-clock time and storage space required for each operation. This is a direct consequence of how general the tool is---it treats the model checkpoint as a generic file and each step simply involves hashing and copying the file. This results in the same checkpoint size every time, regardless of whether it is filled with redundant information. Speeds are similarly consistent, only effected by write contention. Since Git LFS cannot merge models, we report the time and space costs for storing a checkpoint that would be derived from an external model merging tool. Git-Theta, on the other hand, can perform automatic merging via parameter averaging when using the \code{git merge} command.

As expected, Git-Theta is slower because it does more work. Instead of a single read and write to move the checkpoint as a blob, Git-Theta must parse the checkpoint file, process individual parameter groups, and write the modified parameter groups to disk. Git-Theta leverages the embarrassingly parallel nature of parameter processing and makes heavy use of asynchronous and multi-core code. As a result, Git-Theta's performance can also have more variability because it is more sensitive to shared resource utilization.

\begin{figure}[!t]
  \begin{center}
    \centerline{\includegraphics[width=0.8\columnwidth]{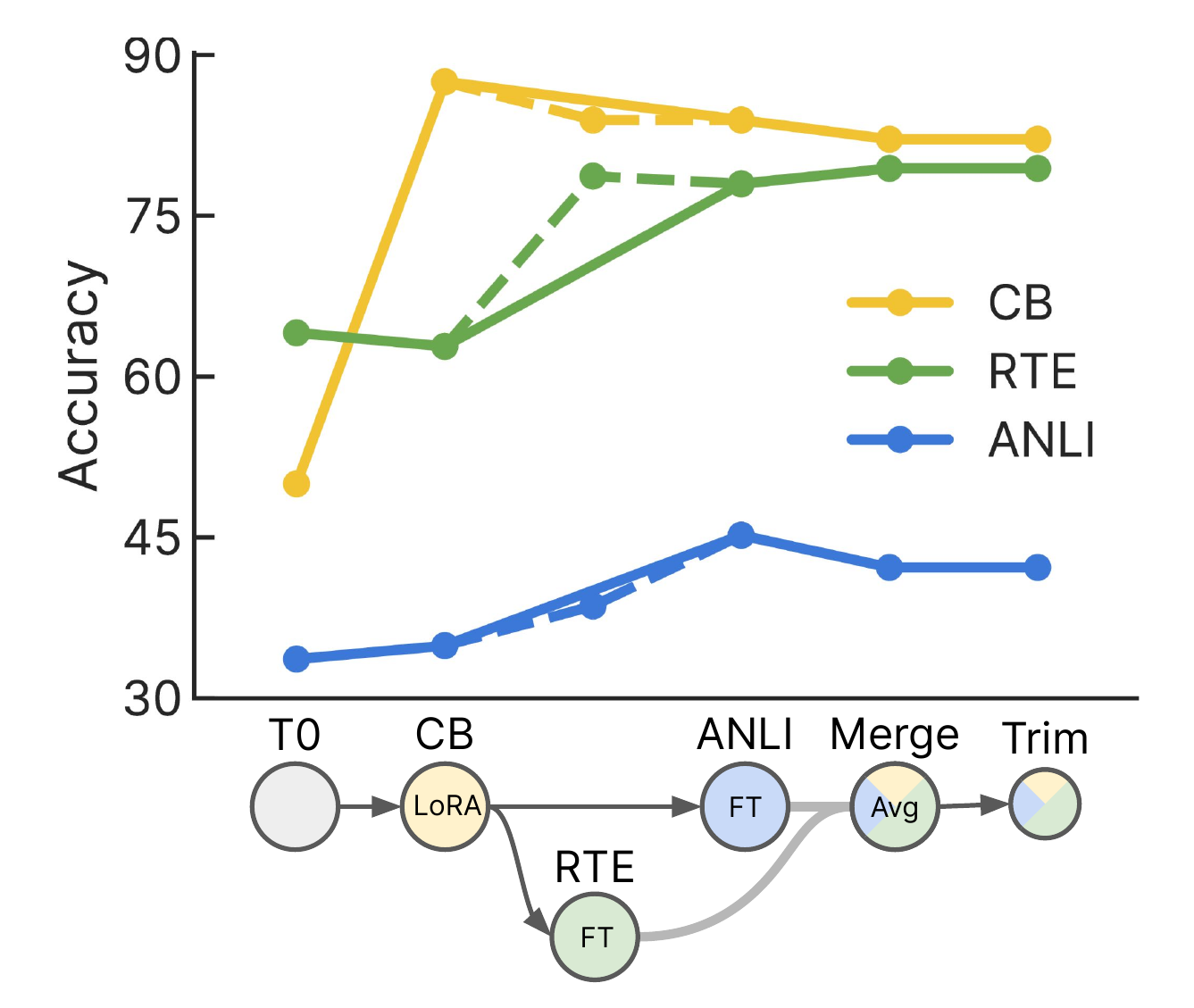}}
    \caption{Model performance at different points in commit history. As expected, merging the RTE and ANLI branches increases performance on RTE.}
    \label{fig:perf}
  \end{center}
  \vskip -0.4in
\end{figure}

Git-Theta's primary advantage is apparent when it comes to storage of non-dense updates. For example, saving the model after training on CB using LoRA required just 0.27 GB to store due to the low-rank nature of the updates. Similarly, the final commit, where sentinels where removed, uses just 1024 kB of storage to track which vocabulary indices were removed. We also see the processing times are reduced when fewer parameters are changed. We note that collaborative model building would likely use a higher proportion of these communication-efficient updates than we have included in our benchmark.

Even for commits that modify the whole model, Git-Theta requires less space due to its use of TensorStore, which compresses the data when it is written to disk. TensorStore's compression is particularly valuable in the first commit since T0 3B was trained using \code{bfloat16} precision but is distributed as a \code{float32} checkpoint. 

\cref{fig:perf} shows the performance at different points in the the commit history. We see that initial training on CB caused a minor degradation in RTE performance but, as suggested in \citet{Choshen2022}, merging an ANLI-trained model with an RTE-trained model increases RTE performance.

\section{Related Work}
\label{sec:related-work}

To the best of our knowledge, all currently available systems for tracking machine learning models (such as DVC\footnote{\url{https://dvc.org/}}, MLFlow \citep{zaharia2018accelerating}, WandB\footnote{\url{https://wandb.ai/}}, Neptune\footnote{\url{https://neptune.ai/}}, and the Hugging Face Hub~\citep{wolf2020transformers}) track a model checkpoint as a single large file (i.e., a blob of data).
As in Git LFS, these systems track large files by using Git to track metadata about the file while keeping the contents in a storage system external to the repository.
Consequently, these systems cannot be used for collaborative model development since they cannot take advantage of recent research on communication-efficient training and model merging.

Like AdapterHub \cite{pfeiffer2020adapterhub}, Git-Theta enables the sharing of cheaply-communicable/storable updates that allow existing pre-trained models to perform a new tasks. Unlike Git-Theta, AdapterHub does not track continual changes to a given model and does not include functionality for merging different versions or histories of a model.

An earlier related system, ModelHub \cite{miao2017modelhub}, aims to track the full lifecycle of a model, including its architecture, hyperparameters, training details, metrics, and more.
ModelHub also includes a query language to find specific models based on the large amount of metadata it stores.
Like Git-Theta, ModelHub includes some functionality for minimizing storage costs, but it still treats the model as a single large tensor of values and does not support merging.
Philosophically, the two systems also differ significantly---ModelHub aims to be a complete system for tracking experiments, models, code, and metadata, whereas Git-Theta is a lightweight Git extension that supports existing workflows and is easily extendible.

Since Git-Theta focuses on tracking changes to a model made by distributed contributors, it is complementary to systems that implement similar functionality for datasets such as Dolt\footnote{\url{https://dolthub.com}}, Pachyderm\footnote{\url{https://pachyderm.com}}, and XetHub~\citep{low2023git}.

\section{Conclusion}\label{sed:conclusion}

Like software before it, machine learning will benefit by moving from a monolithic cathedral of large models built by insular teams to a bazaar of distributed contributors \cite{cat-and-baz1999}. Git-Theta aims to make this transition possible by providing a way to \emph{efficiently} and \emph{meaningfully} track the version history of collaboratively developed machine learning models. 
Future work includes supporting more \code{Checkpoint} types such as T5X \cite{roberts2022t5x}, more \code{Update} types, exposing \code{Serialization} plug-ins to users, and more sophisticated \code{Merge} operations \cite{matena2021merging, ainsworth2022git}. Additionally, we plan to speed up Git-Theta by increasing our use of multi-core parallelism and add easy-to-use per-parameter configuration. We are also working on the ability to directly save a checkpoint into Git-Theta from within a training loop, thus eliminating unnecessary writes when unchanged parameters are checkpointed.

\section*{Acknowledgements}

This material is based upon work supported by the National Science Foundation under Grant No. 2145822. The authors thank Leshem Choshen and Elad Venezian for helpful feedback on Git-Theta's design.

\bibliography{refs}

\begin{thebibliography}{46}
\providecommand{\natexlab}[1]{#1}
\providecommand{\url}[1]{\texttt{#1}}
\expandafter\ifx\csname urlstyle\endcsname\relax
  \providecommand{\doi}[1]{doi: #1}\else
  \providecommand{\doi}{doi: \begingroup \urlstyle{rm}\Url}\fi

\bibitem[Abadi et~al.(2016)Abadi, Barham, Chen, Chen, Davis, Dean, Devin,
  Ghemawat, Irving, Isard, Kudlur, Levenberg, Monga, Moore, Murray, Steiner,
  Tucker, Vasudevan, Warden, Wicke, Yu, and Zheng]{abadi2016tensorflow}
Abadi, M., Barham, P., Chen, J., Chen, Z., Davis, A., Dean, J., Devin, M.,
  Ghemawat, S., Irving, G., Isard, M., Kudlur, M., Levenberg, J., Monga, R.,
  Moore, S., Murray, D.~G., Steiner, B., Tucker, P., Vasudevan, V., Warden, P.,
  Wicke, M., Yu, Y., and Zheng, X.
\newblock {T}ensorflow: {A} {S}ystem for {L}arge-{S}cale {M}achine {L}earning.
\newblock In \emph{12th USENIX Symposium on Operating Systems Design and
  Implementation (OSDI 16)}, pp.\  265--283, 2016.
\newblock URL \url{https://arxiv.org/abs/1605.08695}.

\bibitem[Aghajanyan et~al.(2021)Aghajanyan, Gupta, and
  Zettlemoyer]{aghajanyan2020intrinsic}
Aghajanyan, A., Gupta, S., and Zettlemoyer, L.
\newblock {Intrinsic Dimensionality Explains the Effectiveness of Language
  Model Fine-Tuning}.
\newblock In \emph{Proceedings of the 59th Annual Meeting of the Association
  for Computational Linguistics and the 11th International Joint Conference on
  Natural Language Processing (Volume 1: Long Papers)}, pp.\  7319--7328,
  Online, August 2021. Association for Computational Linguistics.
\newblock \doi{10.18653/v1/2021.acl-long.568}.
\newblock URL \url{https://arxiv.org/abs/2012.13255}.

\bibitem[Ainsworth et~al.(2023)Ainsworth, Hayase, and
  Srinivasa]{ainsworth2022git}
Ainsworth, S., Hayase, J., and Srinivasa, S.
\newblock {Git Re-Basin: Merging Models modulo Permutation Symmetries}.
\newblock In \emph{The Eleventh International Conference on Learning
  Representations}, 2023.
\newblock URL \url{https://arxiv.org/abs/2209.04836}.

\bibitem[Bar-Haim et~al.(2006)Bar-Haim, Dagan, Dolan, Ferro, Giampiccolo,
  Magnini, and Szpektor]{bar2006second}
Bar-Haim, R., Dagan, I., Dolan, B., Ferro, L., Giampiccolo, D., Magnini, B.,
  and Szpektor, I.
\newblock {The Second PASCAL Recognising Textual Entailment Challenge}.
\newblock In \emph{Proceedings of the second PASCAL challenges workshop on
  recognising textual entailment}, volume~6, pp.\  6--4. Venice, 2006.

\bibitem[Ben~Zaken et~al.(2022)Ben~Zaken, Goldberg, and
  Ravfogel]{zaken2021bitfit}
Ben~Zaken, E., Goldberg, Y., and Ravfogel, S.
\newblock {B}it{F}it: Simple parameter-efficient fine-tuning for
  transformer-based masked language-models.
\newblock In \emph{Proceedings of the 60th Annual Meeting of the Association
  for Computational Linguistics (Volume 2: Short Papers)}, pp.\  1--9, Dublin,
  Ireland, May 2022. Association for Computational Linguistics.
\newblock \doi{10.18653/v1/2022.acl-short.1}.
\newblock URL \url{https://arxiv.org/abs/2106.10199}.

\bibitem[Bentivogli et~al.(2009)Bentivogli, Clark, Dagan, and
  Giampiccolo]{bentivogli2009fifth}
Bentivogli, L., Clark, P., Dagan, I., and Giampiccolo, D.
\newblock {The Fifth PASCAL Recognizing Textual Entailment Challenge.}
\newblock In \emph{TAC}, 2009.

\bibitem[Brooks~Jr.(1975)]{brooks1975mythical}
Brooks~Jr., F.~P.
\newblock \emph{{The Mythical Man-Month: Essays on Software Engineering}}.
\newblock Addison-Wesley Publishing Company, Inc., 1975.

\bibitem[Choshen et~al.(2022{\natexlab{a}})Choshen, Venezian, Don-Yehia,
  Slonim, and Katz]{Choshen2022}
Choshen, L., Venezian, E., Don-Yehia, S., Slonim, N., and Katz, Y.
\newblock {Where to start? Analyzing the potential value of intermediate
  models}.
\newblock \emph{arXiv preprint arXiv:2211.00107}, 2022{\natexlab{a}}.
\newblock \doi{10.48550/arxiv.2211.00107}.
\newblock URL \url{https://arxiv.org/abs/2211.00107}.

\bibitem[Choshen et~al.(2022{\natexlab{b}})Choshen, Venezian, Slonim, and
  Katz]{choshen2022fusing}
Choshen, L., Venezian, E., Slonim, N., and Katz, Y.
\newblock Fusing finetuned models for better pretraining.
\newblock \emph{arXiv preprint arXiv:2204.03044}, 2022{\natexlab{b}}.
\newblock URL \url{https://arxiv.org/abs/2204.03044}.

\bibitem[Courbariaux et~al.(2014)Courbariaux, Bengio, and
  David]{Courbariaux2014}
Courbariaux, M., Bengio, Y., and David, J.-P.
\newblock {Training Deep Neural Networks with Low Precision Multiplications}.
\newblock \emph{arXiv preprint arxiv:1412.7024}, 2014.
\newblock URL \url{https://arxiv.org/abs/1412.7024}.

\bibitem[Dagan et~al.(2005)Dagan, Glickman, and Magnini]{dagan2005pascal}
Dagan, I., Glickman, O., and Magnini, B.
\newblock {The PASCAL Recognising Textual Entailment Challenge}.
\newblock In \emph{Machine Learning Challenges Workshop}, pp.\  177--190.
  Springer, 2005.

\bibitem[Datar et~al.(2004)Datar, Immorlica, Indyk, and Mirrokni]{Datar2004}
Datar, M., Immorlica, N., Indyk, P., and Mirrokni, V.~S.
\newblock {Locality-Sensitive Hashing Scheme Based on p-Stable Distributions}.
\newblock In \emph{Proceedings of the Twentieth Annual Symposium on
  Computational Geometry}, SCG '04, pp.\  253–262, New York, NY, USA, 2004.
  Association for Computing Machinery.
\newblock ISBN 1581138857.
\newblock URL \url{https://doi.org/10.1145/997817.997857}.

\bibitem[De~Marneff et~al.(2019)De~Marneff, Simons, and
  Tonhauser]{marneff-simons-tonhauser-2019}
De~Marneff, M.-C., Simons, M., and Tonhauser, J.
\newblock {The CommitmentBank: Investigating projection in naturally occurring
  discourse}.
\newblock In \emph{Proceedings of Sinn und Bedeutung 23}, 2019.

\bibitem[Ding et~al.(2022)Ding, Qin, Yang, Wei, Yang, Su, Hu, Chen, Chan, Chen,
  et~al.]{ding2022delta}
Ding, N., Qin, Y., Yang, G., Wei, F., Yang, Z., Su, Y., Hu, S., Chen, Y., Chan,
  C.-M., Chen, W., et~al.
\newblock Delta tuning: A comprehensive study of parameter efficient methods
  for pre-trained language models.
\newblock \emph{arXiv preprint arXiv:2203.06904}, 2022.
\newblock URL \url{https://arxiv.org/abs/2203.06904}.

\bibitem[Giampiccolo et~al.(2007)Giampiccolo, Magnini, Dagan, and
  Dolan]{giampiccolo2007third}
Giampiccolo, D., Magnini, B., Dagan, I., and Dolan, B.
\newblock {The Third PASCAL Recognizing Textual Entailment Challenge}.
\newblock In \emph{Proceedings of the ACL-PASCAL workshop on textual entailment
  and paraphrasing}, pp.\  1--9. Association for Computational Linguistics,
  2007.

\bibitem[Gionis et~al.(1999)Gionis, Indyk, and Motwani]{gim-99}
Gionis, A., Indyk, P., and Motwani, R.
\newblock {Similarity Search in High Dimensions via Hashing}.
\newblock In \emph{Proceedings of the 25th International Conference on Very
  Large Data Bases}, VLDB '99, pp.\  518–529, San Francisco, CA, USA, 1999.
  Morgan Kaufmann Publishers Inc.
\newblock ISBN 1558606157.

\bibitem[Goldberg(1991)]{goldberg1991every}
Goldberg, D.
\newblock What every computer scientist should know about floating-point
  arithmetic.
\newblock \emph{ACM computing surveys (CSUR)}, 23\penalty0 (1):\penalty0 5--48,
  1991.
\newblock URL \url{https://dl.acm.org/doi/10.1145/103162.103163}.

\bibitem[Guo et~al.(2021)Guo, Rush, and Kim]{guo2020diffpruning}
Guo, D., Rush, A., and Kim, Y.
\newblock Parameter-efficient transfer learning with diff pruning.
\newblock In \emph{Proceedings of the 59th Annual Meeting of the Association
  for Computational Linguistics and the 11th International Joint Conference on
  Natural Language Processing (Volume 1: Long Papers)}, pp.\  4884--4896,
  Online, August 2021. Association for Computational Linguistics.
\newblock \doi{10.18653/v1/2021.acl-long.378}.
\newblock URL \url{https://arxiv.org/abs/2012.07463}.

\bibitem[Gupta et~al.(2015)Gupta, Agrawal, Gopalakrishnan, and
  Narayanan]{Gupta2015}
Gupta, S., Agrawal, A., Gopalakrishnan, K., and Narayanan, P.
\newblock Deep learning with limited numerical precision.
\newblock In \emph{Proceedings of the 32nd International Conference on Machine
  Learning}, volume~37 of \emph{Proceedings of Machine Learning Research}, pp.\
   1737--1746, Lille, France, 07--09 Jul 2015. PMLR.
\newblock URL \url{https://arxiv.org/abs/1502.02551}.

\bibitem[He et~al.(2022)He, Zhou, Ma, Berg-Kirkpatrick, and
  Neubig]{he2022towards}
He, J., Zhou, C., Ma, X., Berg-Kirkpatrick, T., and Neubig, G.
\newblock Towards a unified view of parameter-efficient transfer learning.
\newblock In \emph{International Conference on Learning Representations}, 2022.
\newblock URL \url{https://arxiv.org/abs/2110.04366}.

\bibitem[Heek et~al.(2020)Heek, Levskaya, Oliver, Ritter, Rondepierre, Steiner,
  and van {Z}ee]{flax2020github}
Heek, J., Levskaya, A., Oliver, A., Ritter, M., Rondepierre, B., Steiner, A.,
  and van {Z}ee, M.
\newblock {F}lax: {A} {N}eural {N}etwork {L}ibrary and {E}cosystem for {JAX},
  2020.
\newblock URL \url{http://github.com/google/flax}.

\bibitem[Houlsby et~al.(2019)Houlsby, Giurgiu, Jastrzebski, Morrone,
  De~Laroussilhe, Gesmundo, Attariyan, and Gelly]{houlsby2019adapter}
Houlsby, N., Giurgiu, A., Jastrzebski, S., Morrone, B., De~Laroussilhe, Q.,
  Gesmundo, A., Attariyan, M., and Gelly, S.
\newblock {Parameter-Efficient Transfer Learning for {NLP}}.
\newblock In \emph{Proceedings of the 36th International Conference on Machine
  Learning}, volume~97 of \emph{Proceedings of Machine Learning Research}, pp.\
   2790--2799. PMLR, 09--15 Jun 2019.
\newblock URL \url{https://arxiv.org/abs/1902.00751}.

\bibitem[Hu et~al.(2022)Hu, yelong shen, Wallis, Allen-Zhu, Li, Wang, Wang, and
  Chen]{hu2021lora}
Hu, E.~J., yelong shen, Wallis, P., Allen-Zhu, Z., Li, Y., Wang, S., Wang, L.,
  and Chen, W.
\newblock {Lo{RA}: Low-Rank Adaptation of Large Language Models}.
\newblock In \emph{International Conference on Learning Representations}, 2022.
\newblock URL \url{https://arxiv.org/abs/2106.09685}.

\bibitem[Johnson \& Foote(1988)Johnson and Foote]{johnson1988designing}
Johnson, R.~E. and Foote, B.
\newblock {Designing Reusable Classes}.
\newblock \emph{Journal of Object-Oriented Programming}, 1\penalty0
  (2):\penalty0 22--35, June/July 1988.
\newblock URL \url{http://www.laputan.org/drc.html}.

\bibitem[Lester et~al.(2021)Lester, Al-Rfou, and Constant]{lester2021tuning}
Lester, B., Al-Rfou, R., and Constant, N.
\newblock {The Power of Scale for Parameter-Efficient Prompt Tuning}.
\newblock In \emph{Proceedings of the 2021 Conference on Empirical Methods in
  Natural Language Processing}, pp.\  3045--3059, Online and Punta Cana,
  Dominican Republic, November 2021. Association for Computational Linguistics.
\newblock \doi{10.18653/v1/2021.emnlp-main.243}.
\newblock URL \url{https://arxiv.org/abs/2104.08691}.

\bibitem[Lester et~al.(2022)Lester, Yurtsever, Shakeri, and
  Constant]{lester-etal-2022-recycling}
Lester, B., Yurtsever, J., Shakeri, S., and Constant, N.
\newblock Reducing retraining by recycling parameter-efficient prompts.
\newblock \emph{arXiv preprint arXiv:2208.05577}, aug 2022.
\newblock URL \url{https://arxiv.org/abs/2208.05577}.

\bibitem[Li \& Liang(2021)Li and Liang]{li2021prefix}
Li, X.~L. and Liang, P.
\newblock {Prefix-Tuning: Optimizing Continuous Prompts for Generation}.
\newblock In \emph{Proceedings of the 59th Annual Meeting of the Association
  for Computational Linguistics and the 11th International Joint Conference on
  Natural Language Processing (Volume 1: Long Papers)}, pp.\  4582--4597,
  Online, August 2021. Association for Computational Linguistics.
\newblock \doi{10.18653/v1/2021.acl-long.353}.
\newblock URL \url{https://arxiv.org/abs/2101.00190}.

\bibitem[Liu et~al.(2022)Liu, Tam, Mohammed, Mohta, Huang, Bansal, and
  Raffel]{liu2022tfew}
Liu, H., Tam, D., Mohammed, M., Mohta, J., Huang, T., Bansal, M., and Raffel,
  C.
\newblock {Few-Shot Parameter-Efficient Fine-Tuning is Better and Cheaper than
  In-Context Learning}.
\newblock In \emph{Advances in Neural Information Processing Systems}, 2022.
\newblock URL \url{https://arxiv.org/abs/2205.05638}.

\bibitem[Low et~al.(2023)Low, Arya, Banerjee, Huang, Ronan, Koepke, Godlewski,
  and Nation]{low2023git}
Low, Y., Arya, R., Banerjee, A., Huang, A., Ronan, B., Koepke, H., Godlewski,
  J., and Nation, Z.
\newblock Git is for data.
\newblock In \emph{Conference on Innovative Data Systems Research}, 2023.

\bibitem[Matena \& Raffel(2022)Matena and Raffel]{matena2021merging}
Matena, M.~S. and Raffel, C.
\newblock {Merging Models with Fisher-Weighted Averaging}.
\newblock In \emph{Advances in Neural Information Processing Systems}, 2022.
\newblock URL \url{https://arxiv.org/abs/2111.09832}.

\bibitem[Miao et~al.(2017)Miao, Li, Davis, and Deshpande]{miao2017modelhub}
Miao, H., Li, A., Davis, L.~S., and Deshpande, A.
\newblock {ModelHub}: Deep learning lifecycle management.
\newblock In \emph{2017 IEEE 33rd International Conference on Data Engineering
  (ICDE)}, pp.\  1393--1394. IEEE, 2017.
\newblock URL \url{https://ieeexplore.ieee.org/document/7930088}.

\bibitem[Nie et~al.(2020)Nie, Williams, Dinan, Bansal, Weston, and
  Kiela]{nie2019adversarial}
Nie, Y., Williams, A., Dinan, E., Bansal, M., Weston, J., and Kiela, D.
\newblock {Adversarial {NLI}: A New Benchmark for Natural Language
  Understanding}.
\newblock In \emph{Proceedings of the 58th Annual Meeting of the Association
  for Computational Linguistics}, pp.\  4885--4901, Online, July 2020.
  Association for Computational Linguistics.
\newblock \doi{10.18653/v1/2020.acl-main.441}.
\newblock URL \url{https://arxiv.org/abs/1910.14599}.

\bibitem[Paszke et~al.(2019)Paszke, Gross, Massa, Lerer, Bradbury, Chanan,
  Killeen, Lin, Gimelshein, Antiga, Desmaison, Köpf, Yang, DeVito, Raison,
  Tejani, Chilamkurthy, Steiner, Fang, Bai, and Chintala]{paszke2019pytorch}
Paszke, A., Gross, S., Massa, F., Lerer, A., Bradbury, J., Chanan, G., Killeen,
  T., Lin, Z., Gimelshein, N., Antiga, L., Desmaison, A., Köpf, A., Yang, E.,
  DeVito, Z., Raison, M., Tejani, A., Chilamkurthy, S., Steiner, B., Fang, L.,
  Bai, J., and Chintala, S.
\newblock {PyTorch: An Imperative Style, High-Performance Deep Learning
  Library}.
\newblock In \emph{Proceedings of the 33rd International Conference on Neural
  Information Processing Systems}, 2019.
\newblock URL \url{https://arxiv.org/abs/1912.01703}.

\bibitem[Pfeiffer et~al.(2020)Pfeiffer, R{\"u}ckl{\'e}, Poth, Kamath,
  Vuli{\'c}, Ruder, Cho, and Gurevych]{pfeiffer2020adapterhub}
Pfeiffer, J., R{\"u}ckl{\'e}, A., Poth, C., Kamath, A., Vuli{\'c}, I., Ruder,
  S., Cho, K., and Gurevych, I.
\newblock {AdapterHub}: A framework for adapting transformers.
\newblock In \emph{Proceedings of the 2020 Conference on Empirical Methods in
  Natural Language Processing: System Demonstrations}, pp.\  46--54, 2020.
\newblock URL \url{https://arxiv.org/abs/2007.07779}.

\bibitem[Raffel(2023)]{raffel2023building}
Raffel, C.
\newblock Building machine learning models like open source software.
\newblock \emph{Communications of the ACM}, 66\penalty0 (2):\penalty0 38--40,
  2023.

\bibitem[Raffel et~al.(2020)Raffel, Shazeer, Roberts, Lee, Narang, Matena,
  Zhou, Li, and Liu]{2020t5}
Raffel, C., Shazeer, N., Roberts, A., Lee, K., Narang, S., Matena, M., Zhou,
  Y., Li, W., and Liu, P.~J.
\newblock {Exploring the Limits of Transfer Learning with a Unified
  Text-to-Text Transformer}.
\newblock \emph{Journal of Machine Learning Research}, 21\penalty0
  (140):\penalty0 1--67, 2020.
\newblock URL \url{https://arxiv.org/abs/1910.10683}.

\bibitem[Raymond(1999)]{cat-and-baz1999}
Raymond, E.~S.
\newblock \emph{The Cathedral and the Bazaar}.
\newblock O'Reilly \& Associates, Inc., 1999.
\newblock ISBN 1565927249.

\bibitem[Roberts et~al.(2022)Roberts, Chung, Levskaya, Mishra, Bradbury, Andor,
  Narang, Lester, Gaffney, Mohiuddin, Hawthorne, Lewkowycz, Salcianu, van Zee,
  Austin, Goodman, Soares, Hu, Tsvyashchenko, Chowdhery, Bastings, Bulian,
  Garcia, Ni, Chen, Kenealy, Clark, Lee, Garrette, Lee-Thorp, Raffel, Shazeer,
  Ritter, Bosma, Passos, Maitin-Shepard, Fiedel, Omernick, Saeta, Sepassi,
  Spiridonov, Newlan, and Gesmundo]{roberts2022t5x}
Roberts, A., Chung, H.~W., Levskaya, A., Mishra, G., Bradbury, J., Andor, D.,
  Narang, S., Lester, B., Gaffney, C., Mohiuddin, A., Hawthorne, C., Lewkowycz,
  A., Salcianu, A., van Zee, M., Austin, J., Goodman, S., Soares, L.~B., Hu,
  H., Tsvyashchenko, S., Chowdhery, A., Bastings, J., Bulian, J., Garcia, X.,
  Ni, J., Chen, A., Kenealy, K., Clark, J.~H., Lee, S., Garrette, D.,
  Lee-Thorp, J., Raffel, C., Shazeer, N., Ritter, M., Bosma, M., Passos, A.,
  Maitin-Shepard, J., Fiedel, N., Omernick, M., Saeta, B., Sepassi, R.,
  Spiridonov, A., Newlan, J., and Gesmundo, A.
\newblock Scaling up models and data with $\texttt{t5x}$ and $\texttt{seqio}$.
\newblock \emph{arXiv preprint arXiv:2203.17189}, 2022.
\newblock URL \url{https://arxiv.org/abs/2203.17189}.

\bibitem[Sanh et~al.(2022)Sanh, Webson, Raffel, Bach, Sutawika, Alyafeai,
  Chaffin, Stiegler, Scao, Raja, Dey, Bari, Xu, Thakker, Sharma, Szczechla,
  Kim, Chhablani, Nayak, Datta, Chang, Jiang, Wang, Manica, Shen, Yong, Pandey,
  Bawden, Wang, Neeraj, Rozen, Sharma, Santilli, F{\'{e}}vry, Fries, Teehan,
  Biderman, Gao, Bers, Wolf, and Rush]{sanh2021multitask}
Sanh, V., Webson, A., Raffel, C., Bach, S.~H., Sutawika, L., Alyafeai, Z.,
  Chaffin, A., Stiegler, A., Scao, T.~L., Raja, A., Dey, M., Bari, M.~S., Xu,
  C., Thakker, U., Sharma, S., Szczechla, E., Kim, T., Chhablani, G., Nayak,
  N.~V., Datta, D., Chang, J., Jiang, M.~T., Wang, H., Manica, M., Shen, S.,
  Yong, Z.~X., Pandey, H., Bawden, R., Wang, T., Neeraj, T., Rozen, J., Sharma,
  A., Santilli, A., F{\'{e}}vry, T., Fries, J.~A., Teehan, R., Biderman, S.,
  Gao, L., Bers, T., Wolf, T., and Rush, A.~M.
\newblock {Multitask Prompted Training Enables Zero-Shot Task Generalization}.
\newblock In \emph{International Conference on Learning Representations}, 2022.
\newblock URL \url{https://arxiv.org/abs/2110.08207}.

\bibitem[Sung et~al.(2021)Sung, Nair, and Raffel]{sung2021sparse}
Sung, Y.-L., Nair, V., and Raffel, C.
\newblock {Training Neural Networks with Fixed Sparse Masks}.
\newblock In \emph{Advances in Neural Information Processing Systems}, 2021.
\newblock URL \url{https://arxiv.org/abs/2111.09839}.

\bibitem[TensorStore(2020)]{tensorstore2020}
TensorStore.
\newblock Tensorstore, 2020.
\newblock URL \url{https://github.com/google/tensorstore}.

\bibitem[Van~Durme \& Lall(2010)Van~Durme and Lall]{van-durme-lall-2010-online}
Van~Durme, B. and Lall, A.
\newblock {Online Generation of Locality Sensitive Hash Signatures}.
\newblock In \emph{Proceedings of the {ACL} 2010 Conference Short Papers}, pp.\
   231--235, Uppsala, Sweden, July 2010. Association for Computational
  Linguistics.
\newblock URL \url{https://aclanthology.org/P10-2043}.

\bibitem[Van~Rossum \& Drake~Jr(1995)Van~Rossum and Drake~Jr]{van1995python}
Van~Rossum, G. and Drake~Jr, F.~L.
\newblock \emph{Python Reference Manual}.
\newblock Centrum voor Wiskunde en Informatica Amsterdam, 1995.

\bibitem[Wolf et~al.(2020)Wolf, Debut, Sanh, Chaumond, Delangue, Moi, Cistac,
  Rault, Louf, Funtowicz, et~al.]{wolf2020transformers}
Wolf, T., Debut, L., Sanh, V., Chaumond, J., Delangue, C., Moi, A., Cistac, P.,
  Rault, T., Louf, R., Funtowicz, M., et~al.
\newblock Transformers: State-of-the-art natural language processing.
\newblock In \emph{Proceedings of the 2020 conference on empirical methods in
  natural language processing: system demonstrations}, 2020.

\bibitem[Wortsman et~al.(2022)Wortsman, Ilharco, Kim, Li, Kornblith, Roelofs,
  Lopes, Hajishirzi, Farhadi, Namkoong, and Schmidt]{wortsman2021robust}
Wortsman, M., Ilharco, G., Kim, J.~W., Li, M., Kornblith, S., Roelofs, R.,
  Lopes, R.~G., Hajishirzi, H., Farhadi, A., Namkoong, H., and Schmidt, L.
\newblock {Robust Fine-Tuning of Zero-Shot Models}.
\newblock In \emph{Proceedings of the IEEE/CVF Conference on Computer Vision
  and Pattern Recognition (CVPR)}, pp.\  7959--7971, June 2022.
\newblock URL \url{https://arxiv.org/abs/2109.01903}.

\bibitem[Zaharia et~al.(2018)Zaharia, Chen, Davidson, Ghodsi, Hong, Konwinski,
  Murching, Nykodym, Ogilvie, Parkhe, et~al.]{zaharia2018accelerating}
Zaharia, M., Chen, A., Davidson, A., Ghodsi, A., Hong, S.~A., Konwinski, A.,
  Murching, S., Nykodym, T., Ogilvie, P., Parkhe, M., et~al.
\newblock Accelerating the machine learning lifecycle with {MLflow}.
\newblock \emph{Data Engineering}, pp.\ ~39, 2018.
\newblock URL \url{https://www.usenix.org/conference/opml19/presentation/tut5}.

\end{thebibliography}
\bibliographystyle{icml2023}

\end{document}